# AVOIDING UNDESIRED CHOICES USING INTELLIGENT ADAPTIVE SYSTEMS

Amir Konigsberg

General Motors R&D

## ABSTRACT

*We propose a number of heuristics that can be used for identifying when intransitive choice behaviour is likely to occur in choice situations. We also suggest two methods for avoiding undesired choice behaviour, namely transparent communication and adaptive choice-set generation. We believe that these two ways can contribute to the avoidance of decision biases in choice situations that may often be regretted.*

## KEYWORDS

*Intransitive choice; preference reversals; recommender systems; adaptive systems*

## 1. INTRODUCTION

When faced with a set of items, users sometimes reverse their previous preferences because of the presence of additional items in a choice set, even if those additional items aren't themselves chosen [1] [2] [3]. Because of such sensitivity to the *context* of choice, an item in a choice set can have a *relative* as well as an *absolute* value, and its relative value is often a function of how it relates to other items[1]. Moreover, it is natural to suppose that choices made between items under particular choice-conditions may change under other choice-conditions, even if those choice-conditions don't offer any items that offer tangible gain on a previously considered utility function or decision rule.

In what follows we propose a method for avoiding decision biases that can result from preference reversals caused by inner set dependencies between items in choice sets. Our approach is based on applying previous work on the identification of expected preference reversals resulting from choice intransitivity and inner-set dependencies. Our goal in the present paper is to suggest a number of methods by which an intelligent system can support a user in avoiding decision biases. Our primary focus is decision biases that result from the impact of dependencies between unrelated items in choice sets, but our suggestions can extend to support the avoidance of other kinds of decision biases.

The paper will proceed as follows. In the next section (Section #2) we present a means by which preference reversal behaviour can be modelled [2]. In the course of doing so we present the phenomenon of item dependency in choice situations. Thereafter (Section #3) we generalize the model to include a wider range of choice situations – including various types of choice sets and items. We then turn to a brief discussion of the problem of additivity (Section #4) and follow by pointing to possibilities of a extending our generalized model by incorporating it in a system that learns the values of the conditional utilities of items by monitoring users' selections in

---

[1] See Dan Ariely's now famous *Economist* subscription example. See: [4].

International Journal of Artificial Intelligence & Applications (IJAIA), Vol. 5, No. 2, March 2014

various item spaces (Section #5). We then proceed to outline methods that can be used to support a user in avoiding decision biases such as those that result from inner-set dependencies between items (Section #6).

## 2. A MODEL FOR PREFERENCE REVERSAL BEHAVIOR AND ITEM DEPENDENCY

In order to deal with the interacting options in an option space we define a matrix A, whose $i,j$ entry represents the additional utility, or gain, of the $I^{th}$ item, given the $j^{th}$ item. The entries on the main diagonal of the matrix represent the initial utility of the item, disregarding any other items with which it is presented. Otherwise put: the values in the diagonal represent the utility that item x has in and of itself, independent of the presence of any other item. Utility shall be measured in pure unit-free numbers. The scale is chosen such that the biggest number on the main diagonal is 10, meaning that the largest value that can be attributed to an item is 10. But this is only for demonstrative purposes. In essence, larger or smaller scales can be used.

Additionally, we assume that utilities are additive in this matrix, meaning that as a means of evaluating an item we add together the utilities it gets from every other available item. Hence, in order to find the final utility of an item (in some context or option space), we add up the entries of the item's row of values and the columns of the other available options. Then we compare these sums and determine which item has the highest utility, given the context of the specific option space (note: we are aware that assuming additivity is not trivial. We have chosen to presuppose it since we believe it serves our system best. Nonetheless, we discuss this assumption further below). We also assume that the user whose choices this matrix represents is rational, meaning that he maximizes a utility function by choosing the item, or row, with the highest measure of utility. Let us flesh out this model by way of an illustrative example.

You are on a business trip in an area that you are visiting for the first time, driving a car equipped with a point-of-interest (POI) recommendation system that offers you restaurants, museums, shops, and other services and points of interest upon request or proactively, based on your location and preferences. At some point on your journey the system offers you two options:

1) "Hank's" – a club with live music (H)
2) A local restaurant (R).

You don't have too much information about the local music scene – e.g., what kind of music the local bars play, and you also feel quiet hungry, so you rate the options accordingly – you give Ra 10, and H a 5, reflecting your order of preference, in favor of the restaurant. You now have the incomplete matrix (Table 4):

Table 4

|   | H | R  |
|---|---|----|
| H | 5 |    |
| R |   | 10 |

As can be seen, the utility scores in the matrix are distributed for each item – H and R. Because the items that the system has offered are independent of one another and no information derived from one item can affect the other item, the remaining entries in the matrix are zero (Table 5):

Table 5

|   | H | R | Σ |
|---|---|---|---|
| **H** | 5 | 0 | 5 |
| **R** | 0 | 10 | 10 |

You calculate the sum of all entries in the same row, and, being rational (that is, a maximizer), you decide to go to the restaurant (R), because this item has the greatest overall utility.

Now, before you manage to reveal your choice to the system, the system offers you a third item: a music festival (F). You evaluate this item at 7, and then you also realize that now the item relating to the possibility of going to "Hank's" (H) sounds much better, giving it an additional value of 15, perhaps because the idea of going to a festival (F) made listening to music in a bar (H) more attractive. The important point here is that knowing about item F adds some utility to item H.

You now have the following matrix (table 1):

Table 1

|   | H | R | F | Σ |
|---|---|---|---|---|
| **H** | 5 | 0 | 15 | 20 |
| **R** | 0 | 10 | 0 | 10 |
| **F** | 0 | 0 | 7 | 7 |

As can be seen, the offer of a third item F made you assign a higher utility to the first item (H), and in practice it also changed the result of the inner comparison between the initial items H and R (we can also describe this change on a temporal dimension: whereas at time t1 you preferred item R to item H, at time t2, when F also became a possibility, you preferred H. Your preference thus reversed between t_1 and t_2, as a result of the additional availability of F at t2).

## 3. GENERALIZING THE MODEL

Let us generalize this phenomenon to a case in which one faces n choices, denoted as $c_1,\ldots,c_n$. In such a case A is an n×n matrix, whose elements are $a_{(i,j)}$. As explained above, the main diagonal of the matrix houses the basic independent utilities of the items (Table 2):

Table 2

|   | $c_1$ | $c_2$ | ... | $c_n$ |
|---|---|---|---|---|
| $c_1$ | $U_1$ |   |   |   |
| $c_2$ |   | $U_2$ |   |   |
| ⋮ |   |   | ⋱ |   |
| $c_n$ |   |   |   | $U_n$ |

We proceed to fill in the rest of the matrix by examining the relations between the different items: $a_{(i,j)}$ represents the additional utility of item $c_i$ when we know $c_j$ is present (i.e., when ⟦c⟧_j is also a viable possibility). Note that in most cases, the effect isn't symmetrical, so generally speaking $a_{(i,j)} \neq a_{(j,i)}$. We now get this full matrix (Table 3):

Table 3

|  | $c_1$ | $c_2$ | ... | $c_n$ |
|---|---|---|---|---|
| $c_1$ | $U_1$ | $a_{1,2}$ |  | $a_{1,n}$ |
| $c_2$ | $a_{2,1}$ | $U_2$ |  |  |
| ⋮ |  |  | ⋱ | ⋮ |
| $c_n$ | $a_{n,1}$ |  | ... | $U_n$ |

What we want to suggest is that the total utility of an item in some option-space C is the sum of the relevant row and columns. And we believe that recommender systems can benefit by so calculating an item's utility:

$$U(c_k) = \sum_{i \in C} a_{k,i}$$

Where C is the relevant item space; C is a nonempty subset of the set of all possible items. As can be seen in the formula above, the utility of an item depends on its context, or on the options available besides that item. Understanding the effect of the context over the item's utility allows us to manipulate or predict user preference as a function of the item space C. In the last example, we saw that when C = {H,R}, item R was preferred (i.e. U(R) > U(H)), but when we extended C and made it {H,R,F}, we got U(R) < U(H). By adding more items X,Y,Z, we can (perhaps) change the preference again.

The above equation can be rearranged to get the following result: *the utility of an item in context is the sum of the basic independent utility of that item and the additional utility caused by the presence of all other items in the item space*.

$$U(c_k) = U_k + \sum_{\substack{i \in C \\ i \neq k}} a_{k,i}$$

Henceforth our goal is to suggest a computational method that for every set of items {x_1,...,x_n}, from which a particular choice x_1 is made, is able to determine the conditions under which that choice is reversed (to some different x_k) due to an expected reversal in the user's preferences.

To determine in what context (in what item space) the choice is reversed to x_k - i.e. the conditions under which the total utility of x_k is greater than the total utility of x_1- we consider the difference d=x_k-x_1.

Because choice x_1 is the choice made at the outset, we assume that its utility is greater than x_k's and so d<0. The method we propose tries to reverse the choice by creating conditions within which x_k is of greater value. Hence the method tries to maximize d, or at least make it positive.

The method enables establishing how any other $x_m$ affects both $x_1$ and $x_k$. As a rule, if $x_m$ affects $x_k$ more positively than it affects $x_1$, i.e. if $a_{km} > a_{1m}$ or if it has a positive d: $d_m = a_{km} - a_{1m} > 0$, then adding $x_m$ to our choice space will increase the total utility of $x_k$ relative to $x_1$. By gathering all such $x_m$'s we reach the maximal difference between $x_k$ and $x_1$.

However, this does not yet assure us that $x_k$ will be the selected item. Two additional cases need to be considered for the system to be able to predict the user's preference when other items are at hand:

1. Even with all such $x_m$'s, the total utility of $x_k$ might still be lower than $x_1$'s. Relating to the last example, if the music festival (F) is not my taste, it may add only 3 to C's utility, and then item R (at 10) is still preferred over item H (at 8; Table 6):

Table 6

|   | H | R | F | Σ |
|---|---|---|---|---|
| H | 5 | 0 | 3 | 8 |
| R | 0 | 10 | 0 | 10 |
| F | 0 | 0 | 7 | 7 |

2. Every such $x_m$ affects not only $x_k$ and $x_1$, but other $x_l$'s too. So it is possible that after taking all such $x_m$'s, the total utility of $x_k$ will indeed surpass $x_1$'s, but the total utility of some other item $x_l$ will be even higher. Following the last example, the idea of going to a music festival in a foreign land may sound so attractive to you, that you may prefer it over anything else (Table 7):

Table 7

|   | H | R | F | Σ |
|---|---|---|---|---|
| H | 5 | 0 | 15 | 20 |
| R | 0 | 10 | 0 | 10 |
| F | 0 | 0 | 30 | 30 |

3. If we are looking at larger initial sets and we aim to reverse the choice to specific items in those sets then we may end up with a third item being chosen: for example, if the initial item space was {X,Y,Z}, out of which X was chosen, and in order to change the preference to Y in that set we added T,U,V, we might reach a situation in which the choice in {X,Y,Z,T,U,V} is Z. Hence by adding or subtracting items in the item space a user may proceed to take one of a number of consequent actions:

   a) Sticking with the previous choice;
   b) Reversing to a previously available (but not chosen) option (the one we aimed to reverse the choice to);
   c) Choosing one of the new options;
   d) Reversing the preference such that the option that is chosen is another, different option.

We can now ask several questions about the collection of external items that satisfy the 'positive *d*' condition (by "external items" we mean items that are part of the set of all possible items, but currently not in the presented item space). As a reminder, element m is said to meet the 'positive d' condition if $d_m = a_{km} - a_{1m} > 0$, i.e. if it adds more value to the kth $x_k$ item than it adds to the first item $x_1$.

a. Which external items will give the greatest difference between the two items (the currently chosen option and the one we wish to reverse the option to)? i.e. what options will make me prefer the second item by the largest "gap"? The answer appears to be that all of the items that have a positive *d* – every one of them strengthens the utility of the kth option with respect to the first item, so adding them all up will give the greatest difference.

b. What is the set of basic sufficient combinations of choices (choice spaces) that span the remaining item spaces? Evidently if a certain set of items is sufficient to tip the scales, then any additional item with a positive d will result in reversing the choice (making the difference even greater). Hence, for the collection of all item spaces that give us a total positive d, we may be interested in finding the minimal set of item spaces that spans all other combinations. This set shall be referred to as a "base" to the collection of all combinations that tip the scales. In other words, a set B will be a base if every item space in which the choice is reversed contains an element (i.e. a choice set) of B.

c. What external items will give the minimal yet positive difference? i.e. what items will be sufficient to change one's chosen item (to tip the scales). It should be made clear that the answer to this question is a member of a "base" of the collection of all preference-reversing combinations of items, as it is defined in answer to the previous question.

## 4. A NOTE ABOUT ADDITIVITY

In the system we proposed we assumed that utilities are additive. As a reminder, in order to get the total utility of an item in a choice space, our model sums the initial utility of the item and the additional utilities that it receives from the other items in the space. This method is distinct from other methods such as for instance combining utilities in non-additive ways.

As far as computation time and complexity are concerned, assuming that utility is additive has significant implications. Without this assumption, each and every set of items must be checked independently, since we cannot assume any connection between different item spaces; without assuming additivity, knowing the user chose X out of {X,Y,Z} tells us nothing about the choice set {X,Y,Z,T}. Under the linear assumption, it is sufficient to know the independent effect every item has on any other item, regardless of the current space; in this case it is sufficient to know the entries of the matrix. Hence in a given space, we sum these pairwise effects to get the total effect. Because we assume linearity, we can add up independent partial values to get the value of the whole. If we do not assume additivity, we cannot infer any information about the whole space, even when we have full information about its subsets.

What this means for the model is that with the additivity assumption, we can reduce the amount of item spaces (i.e. subsets of the collection of all items) to be checked from all possible subsets (the number of subsets of a set with n elements is $2^n$), to the number of elements in the matrix ($n^2$). This reduction, from exponential runtime to polynomial, is extremely significant, since exponent grows much faster than any polynomial, and so the computation time of the non-additive method will grow rapidly, and will be impractical for even relatively small values of n. Hence because we assume additivity we assume that the additional gain of a set equals the sum of the gain of the individual elements. We therefore ignore any internal relations within the external group that may affect the total gain on a certain item. By considering every subset ($2^n$) we take these internal relations within the external group into consideration. The effect could indeed be zero; in such a case the two methods will give similar results. But if the effect is nonzero, the "full method" (exponential runtime) is more accurate. For example, the utility of a hamburger with a good bun and quality beef exceeds the arithmetic sum of the utilities of a hamburger with a good bun and a hamburger with quality beef. That is, it could be that one quality $q\_1$ adds some utility $u\_1$ to an item o, and a different quality $q\_2$ adds $u\_2$ to o, yet there is something (an internal relation) in the combination of the two qualities $\{q\_1, q\_2\}$ that gives o a utility greater than $q\_1+q\_2$. The "full method" will capture this; the method proposed here will not. Nonetheless, our system assumes additivity because of its relative simplicity and because the amount of data to be considered under the additivity assumption is significantly smaller, and hence the amount of data to be collected is also much smaller, which means much less for the system to learn.

## 5. LEARNING CONDITIONAL UTILITIES

The system learns the values of the conditional utilities of items (i.e. the entries of the matrix A) by monitoring users' selections in various item spaces. Every selection that a user makes of an item from an item space is translated to an inequality with A's entries, and using many such inequalities, a prediction for the exact values of the matrix may be made. Every selection made in an option space gives N inequalities (N being the number of options in the option space). Large amounts of data give us large amounts of inequalities. Every such inequality poses a constraint; by considering all such constraints, the region (in the $n^2$ space) in which the actual Matrix's entries are to be found is narrowed down. With enough constraints it is therefore possible to give good estimates for the matrix entries.

For instance, if when being presented options 1 and 2 the user chooses 1, then we may assume that $U(c\_1) > U(c\_2)$, i.e. $U\_1+a\_{1,2} > U\_2+a\_{2,1}$. However, if when option 3 is also present the user chooses 2, that means that $U\_1+a\_{1,2}+a\_{1,3} < U\_2+a\_{2,1}+a\_{3,1}$. Every such inequality defines a region in an $n^2$-dimensional world (a dimension for every entry in the matrix). By considering many such regions, we can narrow down the options for every entry, and by that get a good estimation for their exact numerical values.

Several methods may be employed to improve this learning process:

- .Since we do not consider the exact numerical utility, but just compare different utilities, we may normalize all values, such that, for example, $U\_1=1$ (otherwise, we can multiply the whole matrix by a constant to get that, and the recommendation results will be identical. What we mean here is that it doesn't matter if A = [1 2; 3 4] or A = [2 4; 6 8]; all that matters are the proportions. If with the first matrix we arrived at utility 8, the second matrix will give utility 16, and so on, i.e. it's all monotonous (if X > Y with matrix A, this inequality will remain true if we use 2A or 6A). So we can divide all elements of the matrix by a constant and remain with the same properties, so we can assume U1 = a11 = 1 (otherwise, if a11 = 8, we'll just divide it all by a factor of 8.)).

- If the system records many selections made in the same option space, the system may use that information in order to estimate the "gap" between options. For instance, if when given A and B the user chose A 97% of the time, then the gap may be big, but if the user chose A only 51% of the time, then the total utilities of A and B (in that context) are identical, or close to identical. This replaces the inequality with an equality, which is far better computationally, as it reduces the dimension of the problem by one.

- As explained before, the values on A's main diagonal $U\_k=a\_{(k,k)}$ represent the utility the option has in and of itself. Therefore, after a selection has been made, the system may ask to rate the selection in a context free environment, in order to get an estimated value of U_k, or at least an estimation for the effect cause by the context. For example, in book recommendations, out of three books A, B, and C, book B seems very appealing, so you choose to read it and discover that it is in fact boring. You may feel cheated; in its context, the book seemed interesting, however when it was context-free, it was boring. That means that the presence of books A and C had a positive effect on the contextual (total) utility of book B.

## 6. AVOIDING DECISION BIASES

At this stage it is important to make explicit an assumption which we have been making so far. We have been assuming that preference reversals in choice behavior due to dependency between items in choice sets are undesirable. And it is in relation to the undesirability of these situations that we are am proposing methods on how to avoid them. Yet we should also note that it is perfectly clear that there will be cases where preferences shift due to the impact that dependent but non-chosen items have on other items in choice sets and this will be perfectly rational. In such cases items may provide information or ignite emotional considerations that may lead to reasonable and desired shifts in preference, and this seems to be perfectly legitimate from a normative point of view. Moreover, in such cases what is occurring will not be a decision bias, but an informed shift in preference due to new information provided by an additionally present item. There doesn't seem to me to be any single clear cut way of distinguishing between desirable and undesirable shifts in preference that result from inner-set item dependency. But there are a few contending rules of thumb that can be used to identify such undesirable shifts, and these can be used for systems that aim to support users in avoiding undesired choice behaviors:

1. ***Inconsistency with general, or prevalent preferences***: if I always prefer P whenever P and Q are offered together, and in this one instance where T is also on offer I prefer Q, this is an expressive reversal of a dominant preference of mine – namely choosing P when P and Q. From here it seems fair to conclude that if I generally choose P from P and Q, then P is my prevalent preference. Which by choosing Q I am presently going against, and which in turn makes such behavior undesirable.

2. ***Preferences indicative of choice patterns that are later regretted (or retracted)***: if the system learns to identify choices (understood as revealed preferences) that result from item dependency in choice sets which are then retracted, then the system can learn to predict when preference reversals are likely to have undesired consequences, and thus label such choice constellations (particular sets of items) as sensitive to undesirable shifts in preferential behavior.

3. ***Identifying suspect items***: if there are particular items that tend to lead users to change their preferences, as in cases noted in #1 and #2 above, and these items are identified as doing so for many people (statistically viable numbers of people that can be relied on as "collective behavior") then these items may be labeled as "suspect." Choice sets with 'suspect' items may thus be treated with greater caution and choices made from these sets (if different from dominant choice patterns or reflective of retractable behaviors) may be regarded as undesirable with a greater degree of confidence.

Now we turn to the question of how decision biases can be avoided. Or, more specifically, what kind of methods may be used to avoid undesirable choice behaviour. We will propose two methods stemming from the proposed rules of thumb for identifying undesirable preferential shifts mentioned above.

*Transparent communication*: one, quite obvious but generally underappreciated way of helping people avoid decision biases such as intransitive choices, is communicating to them that they are making an intransitive choice. The simplest way of doing so would be choosing the most effective medium or modality by which to say something along the lines of "are you sure you want Q, because you normally choose P when (P and Q)?"

This tactic would seemingly be most useful in cases such as those noted in #1 above, those involving inconsistency with general or prevalent preferences, from which individual instances of choice changes can be differentiated. But transparent communication could also be used in cases such as those mentioned in #2, when choice behaviours that are later regretted or retracted are identified. In cases such as these messages informing the user of past retractable actions might be useful, as would noting the long term benefits (tangible and symbolic) of transitive, or desirable choices, as opposed to choices that may seem appealing in the present choice set, but may be regarded negatively later on.

Transparent communication could also come in useful in cases such as those outlined in #3, where collective user behaviour makes some items more suspect than others in leading to undesired choices. In these situations we might envision highlighting a suspect item, or emphasizing to the user that extra caution ought to be taken when considering which item to choose.

*Adaptive choice-set generation*: if the system expects with a sufficiently high probability that a particular item in a choice set or the specific composition of a choice-set as a whole will lead to undesirable preference reversals, this same system can adapt to the consequence it expects and thus mitigate risk. Using the matrix system proposed above, the system is able not only to detect when item utilities will change such that preferences will be reversed but also what additional items are needed in the set to reverse the reverse choice that the system predicts. So the matrix system can adapt to predicted choice behaviour by generating choice-set with item compositions that accord with the preferences that it believes are of the highest importance to the user. As such, the system's own utility considerations incorporate the predicted utilities of the user and the preferences that it believes the user *really* wants to satisfy.

## CONCLUSION

Building on a method for predicting intransitive choice behaviour resulting from choice-set compositions, we have proposed a number of heuristics that may be used for identifying when intransitive choice behaviour resulting from preference reversals may be regarded as undesirable and thus as something that a supportive system ought to avoid. We have also suggested two methods for avoiding the undesired choice behaviour, namely transparent communication and adaptive choice-set generation. We believe that these two ways can contribute to the avoidance of decision biases that may often be regretted.

**Author**

Amir Konigsberg is a senior scientist at General Motors Research and Development Labs, where he works on advanced technologies in the field of artificial intelligence and human machine interaction. Amir gained his PhD from the Center for the Study of Rationality and Interactive Decision Theory at the Hebrew University in Jerusalem and the Psychology Department at Princeton University.
.